%% file: latex/acl_latex.tex
\pdfoutput=1

\documentclass[11pt]{article}

\usepackage[preprint]{acl}

\usepackage{times}
\usepackage{latexsym}

\usepackage{booktabs}
\usepackage{multirow}
\usepackage{array}
\usepackage{listings}
\usepackage{algorithm}

\usepackage{algpseudocode}

\usepackage{graphicx}
\usepackage{caption}
\usepackage{amsmath}
\usepackage{amssymb}
\usepackage{multirow}   
\usepackage{booktabs}  
\usepackage{amsmath}    
\usepackage{tikz}
\PassOptionsToPackage{table,xcdraw}{xcolor}

\definecolor{Gray}{gray}{0.95}  

\usepackage{xcolor}

\usepackage[T1]{fontenc}

\usepackage[utf8]{inputenc}

\usepackage{microtype}

\usepackage{inconsolata}

\usepackage{graphicx}

\definecolor{teagreen}{HTML}{f3fce6}  
\definecolor{lightblue}{HTML}{ebf9ff}  
\definecolor{front-color}{HTML}{fff9d8}  
\definecolor{lightpink}{HTML}{f7f3fa}  
\usepackage{ulem}

\definecolor{Magenta}{rgb}{0.8, 0.1, 0.6}

\title{DynTok: Dynamic Compression of Visual Tokens for Efficient and Effective Video Understanding}

\author{
	Hongzhi Zhang, 
	Jingyuan Zhang,
        Xingguang Ji,
        Qi Wang,
        Fuzheng Zhang \\
        {Kuaishou Technology, Beijing, China } \\
        \texttt{zhanghongzhi@kuaishou.com}
}

\begin{document}
\maketitle
\begin{abstract}
Typical video modeling methods, such as LLava, represent videos as sequences of visual tokens, which are then processed by the LLM backbone for effective video understanding. However, this approach leads to a massive number of visual tokens, especially for long videos. 
A practical solution is to first extract relevant visual information from the large visual context before feeding it into the LLM backbone, thereby reducing computational overhead. 
In this work, we introduce DynTok, a novel \textbf{Dyn}amic video \textbf{Tok}en compression strategy. 
DynTok adaptively splits visual tokens into groups and merges them within each group, achieving high compression in regions with low information density while preserving essential content.
Our method reduces the number of tokens to 44.4\% of the original size while maintaining comparable performance. 
It further benefits from increasing the number of video frames and achieves 65.3\% on Video-MME and 72.5\% on MLVU.  
By applying this simple yet effective compression method, we expose the redundancy in video token representations and offer insights for designing more efficient video modeling techniques.
\end{abstract}

\section{Introduction}

Multimodal understanding models are rapidly advancing, showcasing remarkable capabilities in image and video understanding. Typically, Vision-Language Models (VLMs)~\cite{Bordes2024AnIT} leverage CLIP to extract visual representations, which are then processed through a connector and finally fed into the large language model (LLM) backbone. 
Compared to image-understanding tasks, video understanding involves processing a large number of video frames.
Efficient encoding of the numerous video frames is the core topic of video understanding. 
For instance, a 10-minute video sampled at 1 frame per second (fps) generates 600 frames. If each frame is modeled with 196 tokens, this results in 117,600 visual tokens being fed into the LLM. For hour-long videos, this can escalate to over 700k context tokens, making comprehension of such long-duration videos exceptionally challenging.
This creates significant obstacles to effectiveness and efficiency. On one hand, extracting relevant information from such extensive contexts is inherently complex. On the other hand, handling ultra-long contexts demands substantial computational resources and memory capacity.

\begin{figure}
    \centering
     \includegraphics[width=0.49\linewidth, height=0.49\linewidth]{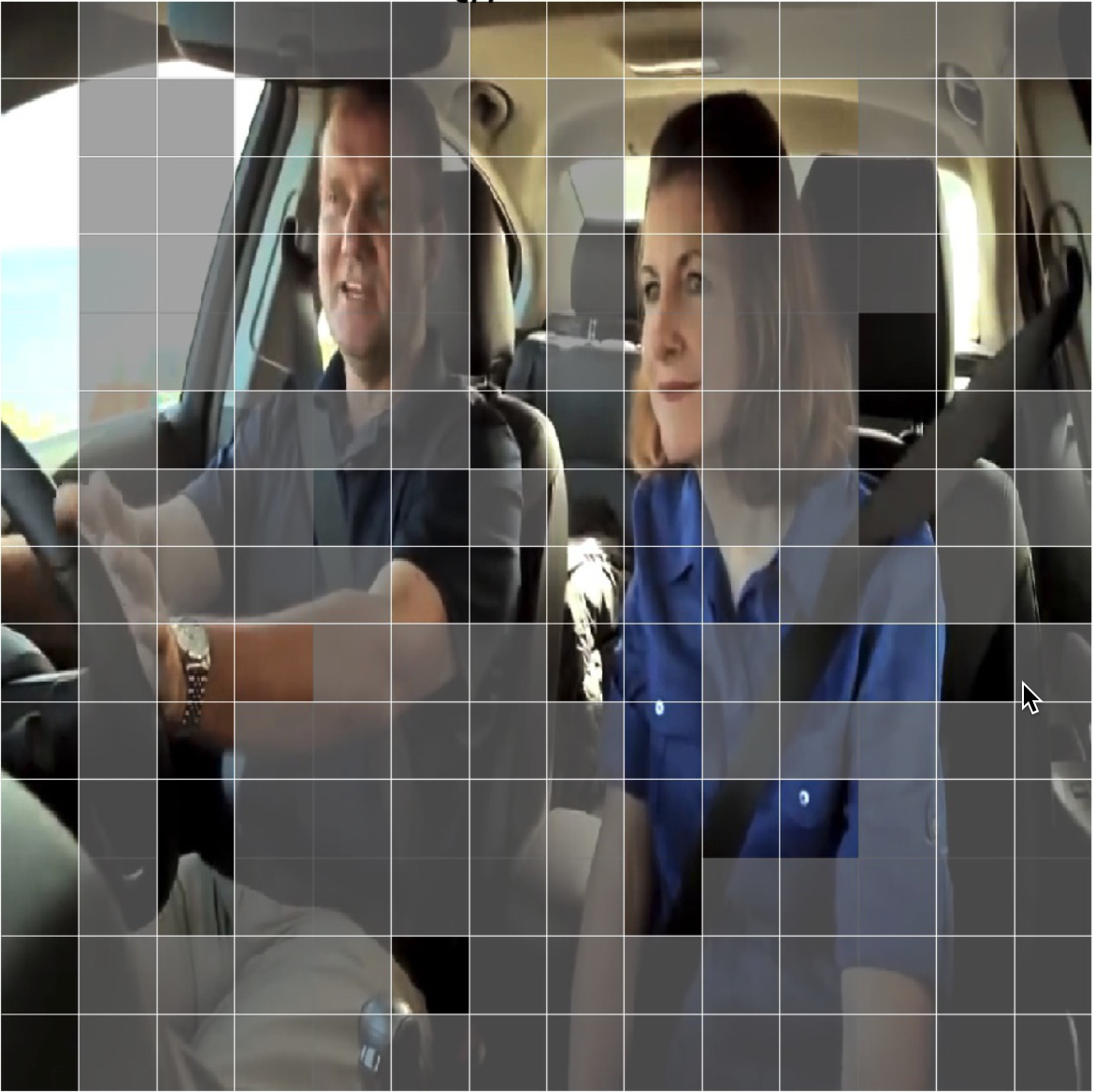}
    \includegraphics[width=0.49\linewidth, height=0.49\linewidth]{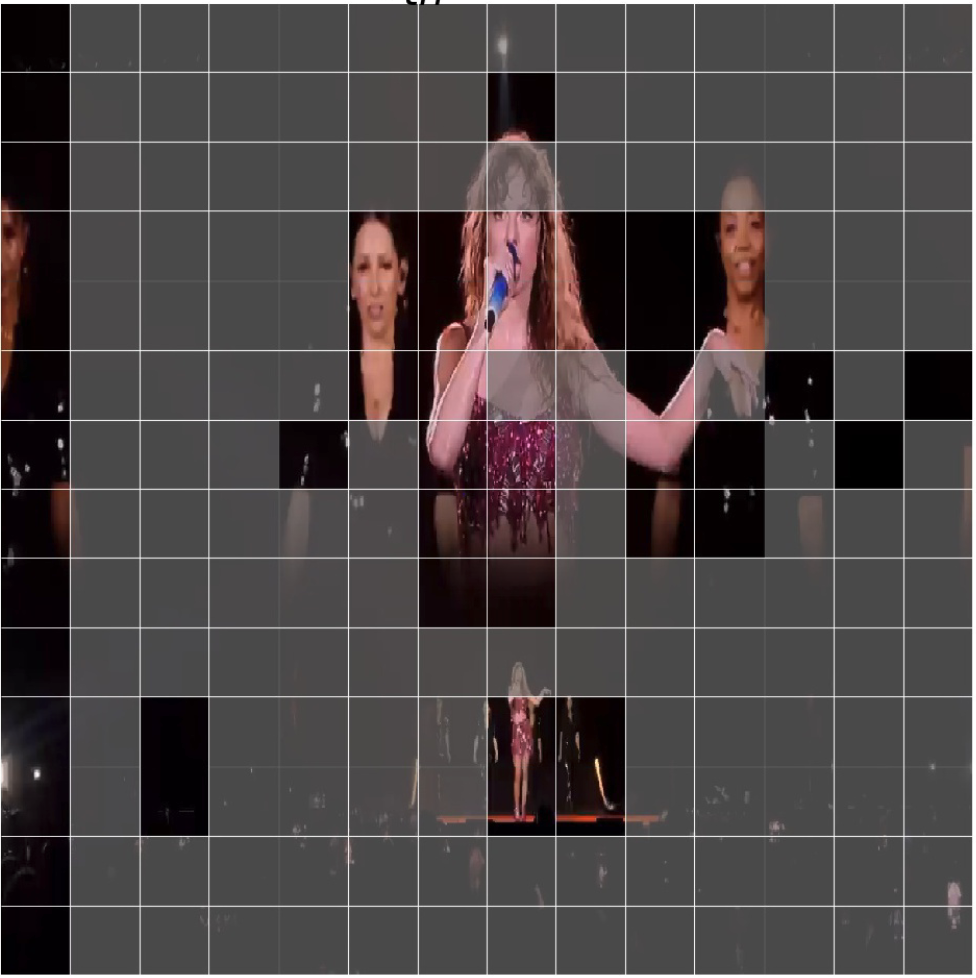}
    \caption{Illustration of two test video frames with their visual token grouping results by DynTok. Each image patch is determined to be masked or not based on its patch similarity to its left neighbor. In this figure, for each patch, if the similarity is greater than the threshold 0.6, it is masked with a gray block. As can be seen, DynTok can effectively retain informative patches with minimal information loss, making it suitable for integration into LLMs for diverse video understanding tasks.
    }
    \label{fig:intro_exmple}
\end{figure}

A number of studies have focused on reducing the number of tokens input into LLMs to fit within the context window and improve computational efficiency. For example, LLaVA~\cite{zhang2025llava} and Dynamic-VLM~\cite{wang2024dynamicvlmsimpledynamicvisual} perform pooling operations on visual tokens, while Q-former~\cite{Li2023BLIP2BL} compresses visual tokens into a fixed-length token sequence~\cite{Bai2023QwenVLAV, Xiao_2021_CVPR}. Some works also select important visual tokens based on attention weights in ViT~\cite{yang2024visionzip} or fuse tokens based on similarities~\cite{Tao2024DyCokeDC}, inspired by~\citet{bolya2022tome}. However, pooling-based methods often fail to account for the varying importance of different tokens, typically using a 2x2 pooling size, with larger pooling sizes leading to performance degradation. Furthermore, current similarity-based token fusion, which is applied across the entire image or video, struggles to preserve spatial information and incurs high computational costs for similarity calculations. In this work, we explore a local token reduction method that more effectively preserves spatial information while minimizing computational overhead.

The main idea of DynTok is illustrated in Figure \ref{fig:intro_exmple}. 
A video frame is typically processed into multiple visual tensors via pretrained visual encoders such as CLIP/SigLIP~\cite{zhai2023sigmoid}, each corresponding to image patches (e.g., 14x14 in LLaVA-OneVision~\cite{li2024llavaonevisioneasyvisualtask} ). It could be observed that the informativeness of these patches varies. 
For example, in the right image, compared to its main subject, the dark background area carries less informational content, indicating the greater potential for token compression. 
Typically, solid color-filled low-density areas exhibit consecutive similar patches, while high information-density regions such as the edges of people or objects differ from preceding tokens. 
Leveraging this phenomenon, we propose to adaptively compress less informative patch tokens, thereby reducing the proportion of low-information-density tokens and decreasing the total number of visual tokens. Furthermore, by merging highly similar tokens within the same row, our method better preserves the spatial relationships of visual tokens while maintaining computational efficiency.

We validated DynTok through extensive experiments on multiple video benchmark datasets. DynTok achieved a slight improvement in model performance while compressing the total number of video tokens to 44.4\% of the baseline method, corresponding to a 2.2x reduction. Moreover, thanks to the compression of tokens per frame, DynTok demonstrates stronger robustness to more extracted frames.  In long video tasks, where more frames are processed, DynTok demonstrates further performance improvements, achieving a VideoMME accuracy of 65.3\%. These results highlight DynTok’s dual benefits: it not only improves modeling efficiency but also reduces the complexity for LLMs to extract information from lengthy sequences of visual tokens.
Our main contribution could be summarized as follows:
\begin{itemize}
    \item We propose a simple yet effective token compression method, DynTok. It achieves high compression ratios in low-information-density patches while effectively preserving critical information by adaptively splitting and fussing the similar adjacent tokens. 
    \item Experiment demonstrates that DynTok reduces video tokens to 44.4\% of the baseline (a 2.2x reduction) while maintaining performance. DynTok could benefit more from the increasing of video frames, enhancing long video understanding by 1.5\% on MLVU and 1.7\% on Video-MME while using a similar number of tokens compared to the baseline. 
\end{itemize}

\section{Related works}

\subsection{Visual language model for video understanding}

VLMs have achieved significant milestones in both image-text integration~\cite{openai2023gpt4v, openai2024gpt4o, liu2023visual, alayrac2022flamingo, tong2024cambrian, Li2023BLIP2BL} and video comprehension~\cite{geminiteam2024gemini15unlockingmultimodal, li2024llavaonevisioneasyvisualtask, zhang2024videoinstructiontuningsynthetic, wang2024qwen2vlenhancingvisionlanguagemodels}. Unlike image understanding, video modeling requires processing sequences of images and capturing the dynamic relationships between frames.
Recent works, such as MiraData~\cite{ju2025miradata} and LLaVA-Video~\cite{zhang2024videoinstructiontuningsynthetic}, focus on leveraging existing VLMs to generate high-quality video captioning or question-answering datasets. 
Meanwhile, models like VideoChat~\cite{li2023videochat}, VideoLLaMa2~\cite{cheng2024videollama}, and Aria~\cite{li2024aria} explore video encoding modules, methods for integrating video encoders with large language models (LLMs), and modifications to LLM architectures. Additionally, there has been significant work on handling long video contexts~\cite{weng2024longvlm, xue2024longvila}.

However, many of these approaches rely on fixed token sizes for distinct frames to represent video, leading to redundant and lengthy video inputs. This paper addresses the challenge of reducing redundancy in video data through spatial compression, aiming to mitigate model overload during video processing.

\subsection{Visual token compression}
Video inputs are typically represented as separate frames, resulting in a substantial increase in the number of visual tokens as the video lengthens. Existing video token compression methods can be broadly classified into two categories: strategy-based and learning-based approaches.

Strategy-based methods focus on dynamically resizing or merging visual tokens. For example, LLaVA-Video~\cite{zhang2024video} divides frames into high-resolution parts for detailed information and low-resolution parts for dynamic content. Dynamic-VLM compresses similar visual tokens within each frame, while FrameFusion~\cite{fu2024framefusioncombiningsimilarityimportance} employs temporal merging and spatial pruning in a two-stage compression process. Additionally, methods like LLaVA-PruMerge~\cite{shang2024llava} and VisionZip~\cite{yang2024visionzip} leverage attention distributions to guide token selection.
While these approaches effectively reduce token redundancy, they may struggle to accurately capture the spatial and temporal distribution of visual tokens, potentially compromising the model’s ability to understand the full context of the video.

Learning-based methods, in contrast, train compressors to directly compress the tokens corresponding to each frame, learning optimal compression patterns through model training. For instance, MobileVLM~\cite{chu2023mobilevlm}, TokenPacker~\cite{li2024tokenpacker}, MQT~\cite{hu2024matryoshka}, and MiniCPM-V~\cite{yao2024minicpmvgpt4vlevelmllm} reduce the number of visual tokens per frame using purpose-built downsampling projectors. Blip3-Video~\cite{ryoo2024xgen} and InternVideo2~\cite{wang2024internvideo2} employ temporal encoder layers to aggregate and compress visual tokens over time. Similarly, VideoChat~\cite{li2023videochat} and LLaVA-mini~\cite{zhang2025llava} use multi-stage fusion strategies to further compress video inputs.
However, many of these methods apply uniform compression, reducing each frame to a fixed number of tokens. These approaches overlook the dynamic nature of visual redundancy, where the amount of compression needed can vary across frames based on content complexity and relevance. As a result, these methods may not fully exploit the potential for more efficient token compression in videos.

\section{Method}
\subsection{Preliminary}
Given one or more videos, denoted as $V$, and a question $q$, the goal of video understanding is to generate the corresponding answer $a$ based on the content of the video. 
We adopt the LLaVA architecture~\cite{li2024llavaonevisioneasyvisualtask}, which effectively integrates the pretrained LLM with visual inputs. The architecture consists of three main components: a vision encoder, an LLM backbone, and an MLP connector. 
\textcolor{black}{The vision encoder processes video frames to extract visual representations. These representations are then mapped to visual tokens in the word embedding space of the LLM through the MLP connector. Finally, the LLM leverages the visual information to fully comprehend the query and generate a response. } 
As mentioned in the introduction, the token sequence for video understanding could be very long and the computational complexity of an LLM can be quadratic with respect to the sequence length. Moreover, the visual tokens account for the majority.
DynTok processed the visual tokens before they were fed into the LLM. 

Formally, suppose $t$ frames are extracted from the input video. The visual encoder encodes the frames into a tensor $X$ with shape $(t,h,w,d_{clip})$, where $h$ and $w$ are the number of patches in the horizontal and vertical directions respectively, and $d_{clip}$ is the representation dimension. Then the MLP layer transforms the representation into the corresponding visual tokens $H \in \mathbb{R}^{t, h, w, d_{emb}}$, where $d_{emb}$ is the dimension of the LLM embedding layer.
\begin{figure*}
    \centering
    \includegraphics[width=1\linewidth]{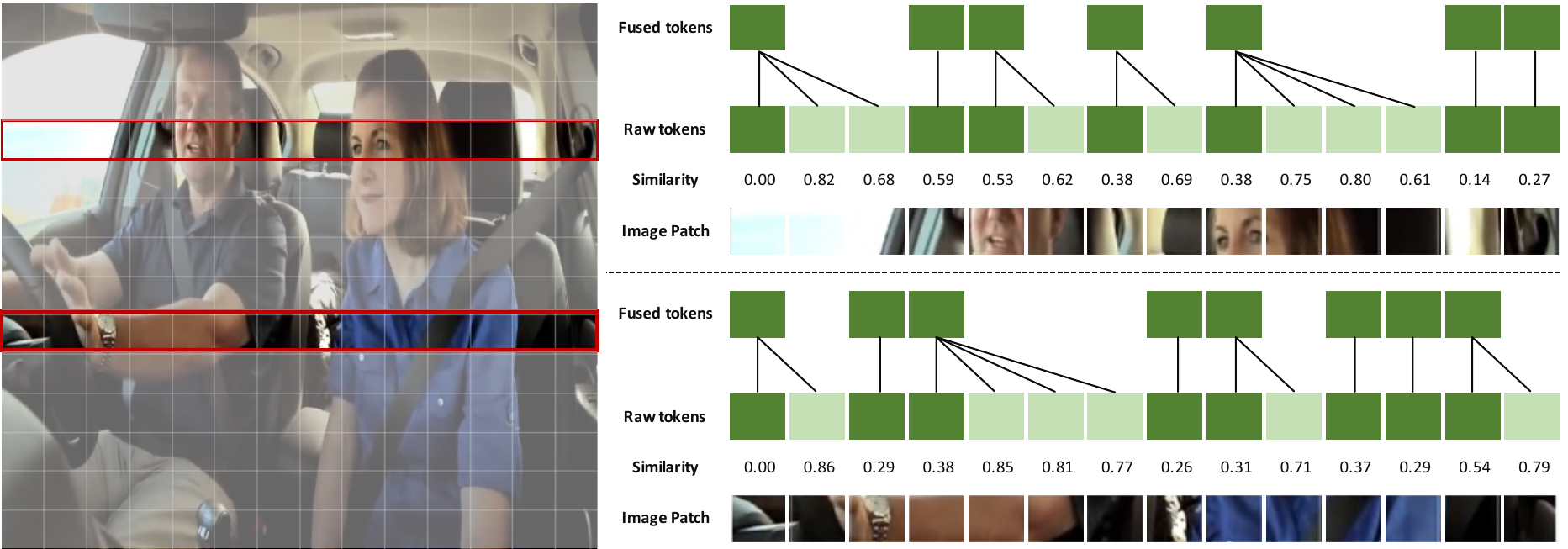}
    \caption{Illustration of the proposed token compression method, DynTok. For each row of image patches, every visual token of a patch is compared with the one corresponding to its left patch. If the similarity exceeds a predefined threshold (0.6 in this example), the token merges into the preceding group; otherwise, it initiates a new group as a primary token. Tokens in the same group are fused together to reduce the number of visual tokens finally.} 
    \label{fig:method}
\end{figure*}

\subsection{Token level compression}
\label{sec:dyntok_method}
Typically, $H$ will be flatten into the visual token tensor with size ${(t \times h \times w, d_{emb}})$ and then fed into the LLM. 
Instead of adopting all tokens, DynTok compresses the visual tokens into $H'\in \mathbb{R}^{l,d_{emb}}$ based on the visual representation $X$, where $l<t\times h \times w$ is the number of reduced token. 

For each video frame represented into a tensor of size $(h,w,d_{emb})$, i.e. there are $(h,w)$ tensors corresponding to $(h,w)$ image patches, 
the main idea of DynTok is to merge adjacent visual tokens within each row. We first describe the specific method for calculating token similarity, followed by the process of dynamically constructing token groups and implementing dynamic information compression.

\paragraph{Token similarity measurement} Each image patch is represented as  $X_{i,j,k}\in\mathbf{R}^{d_{clip}}$, and then transformed into a visual token $H_{i,j,k}\in\mathbf{R}^{d_{emb}}$. 
We simplify the notation to $x_{k}\in\mathbf{R}^{d_{clip}}$ and  $h_{k}\in\mathbf{R}^{d_{emb}}$, by omitting the first two dimensions here.
The similarity between two adjacent visual tokens is computed based on the CLIP representation tensor:
\[s_{(k-1,k)}=\frac{x_{k-1} \cdot x_k}{\| x_{k-1} \| \| x_k \|}.\]
The CLIP representation is used here, by considering that the CLIP or SigLIP trained with contrastive learning cosine loss works better with cosine similarity than with the visual tokens in the embedding space~\cite{zhou-etal-2022-problems}.

\paragraph{Dynamic token merging}

It is interesting to observe that adjacent image patches with solid-color blocks, typically have a high similarity with adjacent nodes in Figure \ref{fig:intro_exmple}. 
DynTok aims to fuse the adjacent image patches to reduce the number of visual tokens while keep the most information. 

For each row of image patches $(h_0,h_1, ... ,h_{w-1} )\in \mathbb{R}^{(w,d_{emb})}$, these patches are split into groups based on a similarity threshold hyperparameter $S_{th}$. 
Initially, the first token patch forms a new token group. For each subsequent visual token $h_{k}$, if the similarity $s_{(k-1, k)}$ between this token and its predecessor exceeds the threshold, it indicates the token carries little new information. In this case, $h_{k}$ is added to the token group containing $h_{k-1}$. 
If the similarity does not surpass the threshold $S_{th}$, it suggests a significant content difference, and thus a new token group is created starting from $h_{k}$. The visual token groups could be formally denoted as $[(h_s,h_{s+1},.., h_e)_{h'}] $, and we denote the number of groups as $h'$. $h'\leq h$ is a dynamic value depending on the similarity between adjacent tokens.

At the fusing stage, the visual tokens within each token group are averaged to form a new visual token tensor of size $(l',d_{emb})$. 
However, the number of tokens at each row varies, making it difficult for the LLM to determine the spatial information of each token, e.g. LLM is not informed where a new row starts. 
Therefore, we add a grid marker at the end of each row to keep the spatial information.  
Finally, the visual tokens from different rows are concatenated together as $H'\in \mathbb{R}^{l,d_{emb}}$ .

In contrast to static pooling methods like bilinear pooling \cite{wang2024dynamicvlmsimpledynamicvisual}, we dynamically construct token groups based on token similarity. 
By fusing tokens with low information density, we can maintain a good modeling performance with fewer tokens.
DynTok, as a local token merge strategy, can better preserve spatial information, and it only requires calculating the similarity between the current token and its preceding token. The computational complexity has a linear relationship with the token length.

\subsection{Model Training}
DynTok is a parameter-free token compression method that merges visual tokens carrying similar information. As a result, DynTok can be applied in a zero-shot manner to existing models. However, using in the zero-shot manner may encounter a training-inference mismatch issue. To mitigate this, the model can be trained according to the DynTok configuration to become familiar with the compressed tokens.

\input{latex/main_result_table}

\section{Experiments}

\subsection{Training Configurations and Evaluations}
\paragraph{Training data.}
The training data is divided into two parts: single-image and video data.
Firstly, we train the model with single-image data to acquire basic visual understanding, the trianing data is identical to the LLAVA-OneVision\cite{li2024llavaonevisioneasyvisualtask}  single-image stage. 
Our Video dataset consists of LLAVA-Video~\cite{zhang2024videoinstructiontuningsynthetic}, VideoInstruct~\cite{Maaz2023VideoChatGPT}, VCG-Plus~\cite{maaz2024videogpt+}, and diverse video question answering/classification data from the training mixture of~\cite{li2024mvbench}.
We remove videos that could not be downloaded or opened, resulting in a final dataset of  1.79M  training samples for the video stage.

\paragraph{Model configurations and hyper-parameters settings.} 
We first drive a model with single-image understanding ability as the warm-up stage before our video training stage.
We adopt the model architecture, training procedure and dataset following LLaVA-OneVision~\cite{li2024llavaonevisioneasyvisualtask}, except we utilize Qwen2.5-7B-instruct~\cite{qwen2.5} as our LLM Backbone.
We do not apply dynamic token compression during this stage.
After that, we train the video understanding model mostly following the experimental setting of LLaVA-Video~\cite{zhang2024videoinstructiontuningsynthetic}.
We first resize each frame to a fixed size of 378$\times$378 and leverage SigLIP to encode this new image, resulting with an visual matrix of 28$\times$28.
The bilinear pooling of stride 2 is performed to reduce the grid into 14$\times$14, i.e. there are 196 tokens for each frame. 
And we apply DynTok on these visual tokens.

In the training, we randomly set the value $S_{th}$ from a set $\{0.4, 0.45, 0.5, 0.55, 0.6\}$ in order to support different compression ratio. Learning rate of the MLP adapter and LLM backbone is set to 1e-5, and 2e-6 for the ViT tower.   
As mentioned in Section~\ref{sec:dyntok_method}, a grid marker is added to the end of each row in order to keep the spatial information. 
The model is trained for 1 epoch on our dataset with the global batch-size 512. 
For a fair comparison, we train a baseline model using identical setting without the DynTok proposal.      
The experiment is conducted on 16 nodes with NVIDIA 8xH100 GPUs. With DeepSpeed Zero2 optimization, the video training stage taking approximately 16 hours.

\paragraph{Evaluation.} 

We evaluate our model on six benchmarks: MVBench~\cite{li2024mvbench}, PerceptionTest~\cite{patraucean2023perception}, NextQA~\cite{Xiao_2021_CVPR}, LongVideoBench~\cite{wu2024longvideobench}, MLVU~\cite{zhou2024mlvu} and Video-MME~\cite{fu2024video}, covering both the perception  and reasoning tasks on videos of various durations. 
MVBench, PerceptionTest and NextQA consist of  relative short videos, and the average duration are 16, 23 and 44 seconds respectively. 
LongVideoBench and MLVU focus on long videos, with the average video duration of 473 and 651 seconds.
Video-MME is a comprehensive video understanding benchmark and the videos are spited into short, medium and long with 0\textasciitilde 2, 4\textasciitilde 15 and 30\textasciitilde60 minutes respectively.
The evaluation is performed by utilizing LMMs-Eval~\cite{zhang2024lmmsevalrealitycheckevaluation} with the default configuration as LLaVA-Video and LLaVA-OneVision.

\input{latex/videomme_category_result_table}

\subsection{Main Results}
Results are listed in Table \ref{tab:main_result}. 
First of all, our baseline model, trained on 10 million samples, achieves performance close to leading open-source models across a range of benchmarks. 
Building upon the collected dataset, we establish a strong comparative baseline for our model. We would make this dataset open avaliable, provide the community with a lightweight near SOTA training dataset. 
The baseline model typically get best performance on short video tasks with max number of frames set to 64 for the first three benchmarks with short video, and 96 frames for the long video benchmarks.

In Zero-shot DynTok, we apply the parameter-free DynTok method directly to the baseline model with $S_{th}$ set to 0.6, and keep other settings unchanged.
Without training on any fused tokens, DynTok$zero$ reduces the number of visual tokens over 2$\times$, while preserving accuracy above 98\% on PerceptionTest, NextQA, LongVideo, MLVU, and Video-MME, and 96.0\% on MVbench. 
The result indicates that DynTok effectively keeps the most informative visual tokens. The impact on different tasks would be covered in Section~\ref{subsection:depth_analyssis}. 

With the integration of the DynTok strategy during model training, we achieve a more than 2$\times$ reduction in visual tokens.
Unlike DynTok-$zero$ which experiences a slight drop in accuracy, DynTok surpasses the baseline model while maintaining the same number of input video frames. 
By consolidating similar visual tokens, the computational cost can be significantly reduced. Meanwhile, the DynTok strategy simplifies information retrieval from the long context of visual tokens, thereby even slightly enhancing model performance.

It is also observed that the DynTok model could benefit from more input frames on long video understanding compared to the baseline. 
The results are listed in the last row of Table~\ref{tab:main_result} and more details are shown in Figure~\ref{fig:frame_impact}. DynTok achieves 65.3\% accuracy on Video-MME with 160 frames extracted, resulting in 15k tokens. This is roughly equivalent to the baseline's token consumption when using 64 frames.

\subsection{In-Depth Analysis}
\label{subsection:depth_analyssis}

\paragraph{Different compression ratios.} 

\begin{figure}[h]
    \centering
    \includegraphics[width=1.08\linewidth]{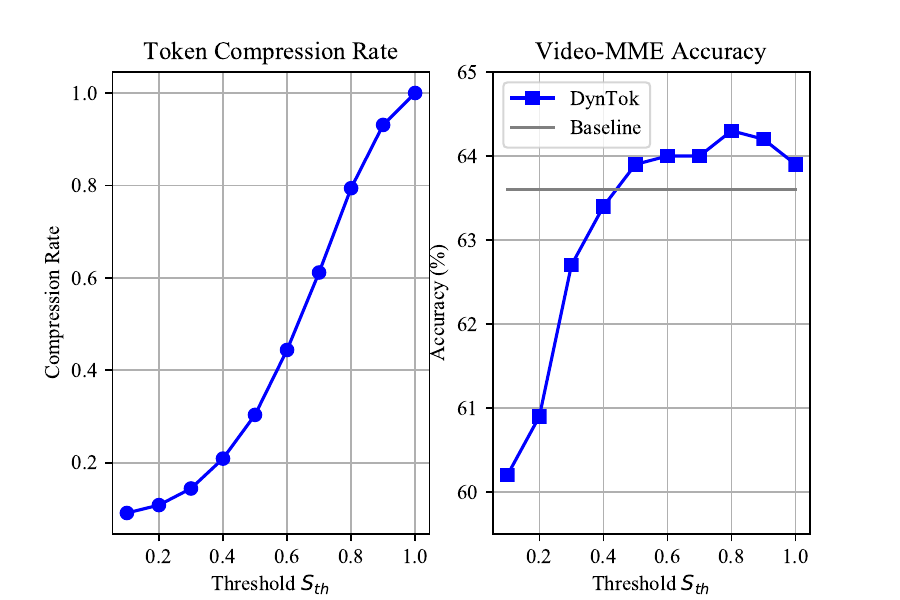}
    \caption{Results of varying the token merging threshold $S_{th}$ on the visual token compression ratio and the Video-MME accuracy. The frame number of the baseline and DynTok is set to 96. }
    \label{fig:compression_ratio}
\end{figure}

\begin{figure*}[t]
    \centering
    \includegraphics[width=1\linewidth]{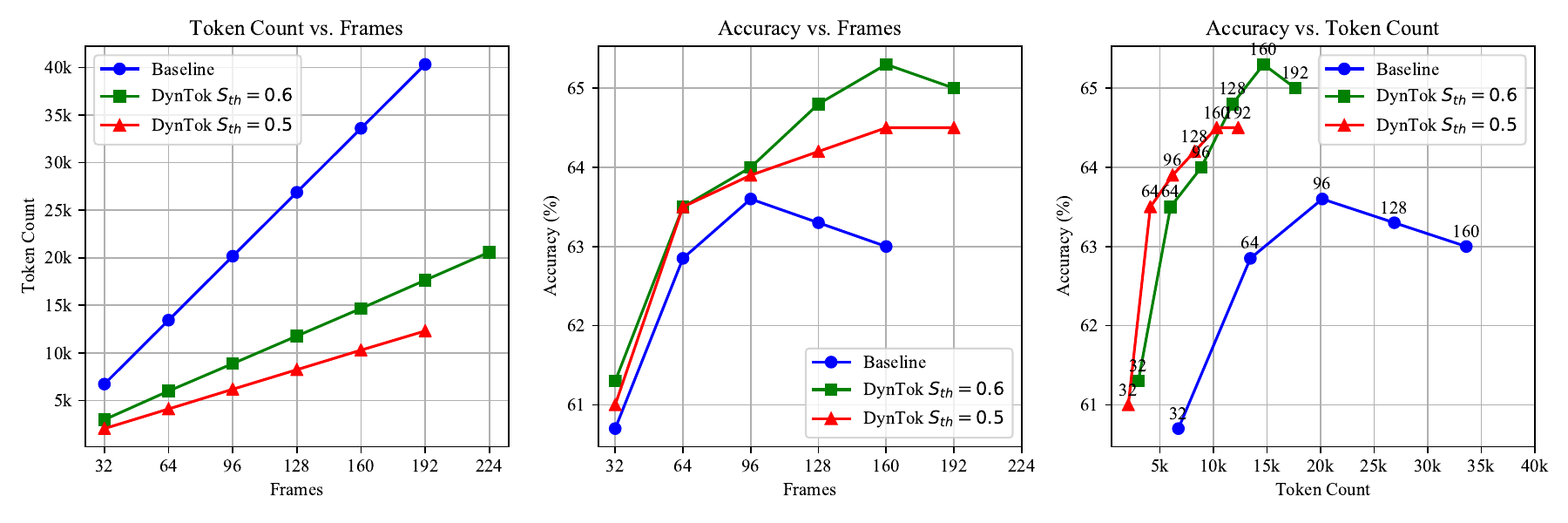}
    \caption{Model performance on Video-MME by varying numbers of video frames used.}
    \label{fig:frame_impact}
\end{figure*}

\begin{figure*}[t]
    \centering
    \includegraphics[width=1\linewidth]{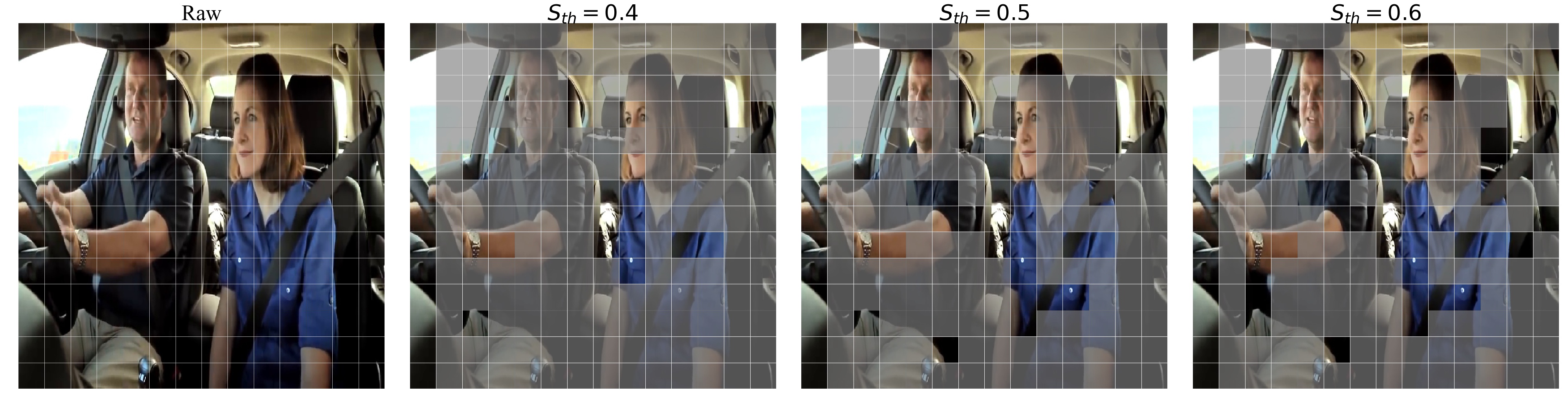}
    \caption{The impact of varying the token merging threshold on token grouping results for the same test image.}
    \label{fig:case_study1}
\end{figure*}

\begin{figure*}[t]
    \centering
    \includegraphics[width=1\linewidth]{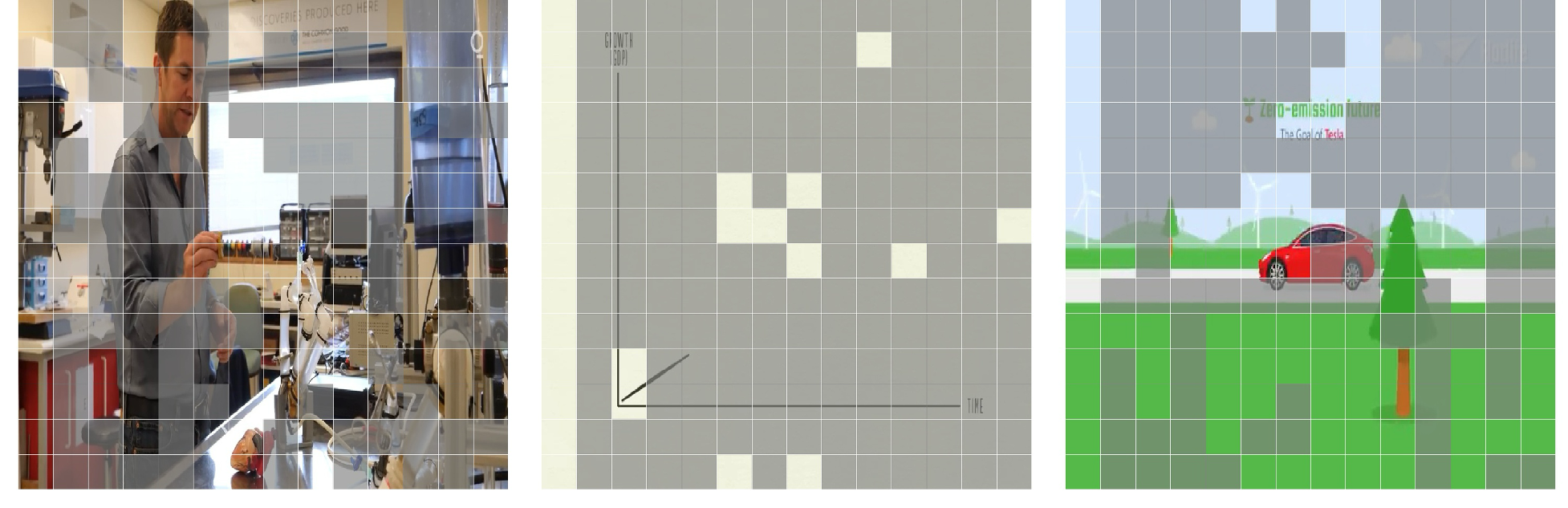}
    \caption{The token grouping results on different test images. The token merging threshold is set as 0.6.}
    \label{fig:case_study2}
\end{figure*}

Figure~\ref{fig:compression_ratio} shows the visual token compression ratio and the corresponding accuracy of Video-MME under different similarity threshold $S_{th}$'s. 
As expected, a lower threshold leads to higher compression of visual tokens.
With the threshold $S_{th}$ set to 0.4, 0.5, and 0.6, the number of visual tokens is reduced to 20\%, 30\%, and 44.4\%, respectively, corresponding to 5$\times$, 3$\times$, and 2.3$\times$ compression. When $S_{th}$ is set to $0.4$, despite a 5$\times$ compression, i.e. the total token number 20.1k is reduced to approximately 4k,  DynTok performs comparably to the baseline (63.4\% vs. 63.6\%). Increasing $S_{th}$ to 0.5 achieves a 3$\times$ compression while surpassing the baseline accuracy. Further increasing $S_{th}$ initially leads to a slight improvement, followed by a minor decline.
Overall, setting $S_{th}$ at the range of 0.4 to 0.6 provides an optimal balance between efficiency and the model performance.

\paragraph{Increasing the number of frames.} 
The model performance with different frames on the baseline and DynTok are shown in Figure~\ref{fig:frame_impact}. 
Overall, DynTok shows stable performance improvement as the number of frames increases.
DynTok introduces a constant compression ratio on the number of visual tokens and outperforms the baseline across different numbers of extracted frames.
While the baseline performance declines with 128 frames or more, DynTok continues to show improvements as the number of frames increases to 128 and 160. 
These results indicate that DynTok effectively preserves visual knowledge while simultaneously reducing the number of tokens, thereby easing the difficulty for the LLM to extract information from long visual token sequences. 

Finally, the right subplot shows the number of visual tokens and the accuracy. DynTok consistently outperforms the baseline on different visual token budgets. 
The results analysis also helps guide reasonable parameter configurations across different token budgets. For token budgets below 10k, setting $S_{th}$ to 0.5 yields better performance than 0.6 with more video frames. 
With a higher token budget, it is preferable to increase the number of extracted frames while slightly decreasing the compression ratio. 
Essentially, this is a balance between increasing the number of observed frames and the compression ratio per frame.
\paragraph{Impact on different tasks.}

We analyze the impact of DynTok on different kinds of tasks on Video-MME, and the results are listed in Table~\ref{tab:model_comparison_on_different_tasks}. 
For DynTok$zero$, performance declines are most noticeable in OCR and action recognition/reasoning tasks. Only slight improvements are observed in temporal perception and reasoning tasks. 
Performance on the rest tasks remains generally stable.
For OCR and action recognition tasks, directly merging tokens  may lead to the confusing on fine-grained details, thereby harming the model's recognition capability. However, for temporal perception and reasoning tasks, which rely  on cross-frame information fusion, reducing the token count per frame enhances the model's ability to capture cross-frame information.

After training with the DynTok strategy, performance of OCR, action recognition, and reasoning tasks either matches or surpasses the baseline. Meanwhile, the advantage in temporal perception tasks is further enhanced. 
Overall, the model adapts well to the token compression scenario, achieving performance improvement over the baseline with only 44.4\% of the visual tokens.

\paragraph{Case study.} 
We analyze the similarity between image patches and visualize the start patch of the visual token grouping in Figure~\ref{fig:case_study1} and Figure~\ref{fig:case_study2} (as well as Figure~\ref{fig:intro_exmple}). 
The token grouping results of DynTok follow a horizontal structure, where the first visual token in each row always initiates a new group.
Tokens with high consistency with preceding tokens are split into the same group on the left and shaded in gray in the figure. 

Figure~\ref{fig:case_study1} presents the results for the same image under different threshold $S_{th}$ values. 
When $S_{th}$ is set to 0.4, most patches in each row are assigned to one group. At a threshold of 0.5, key regions in the image, such as the woman's face and the man's watch, form new visual token groups. When $S_{th}$ is raised to 0.6, important visual patches are better preserved, while background regions, such as the lathe’s color and the dark area in the lower right corner, undergo greater compression.

We now investigate the results across different test cases.  The black background behind the singer in the right image of Figure~\ref{fig:intro_exmple}, the window in middle of the left image in Figure~\ref{fig:case_study2}, and the solid-colored filled areas in the last two images in Figure~\ref{fig:case_study2} are integrated into larger token groups, leading to higher compression. In contrast, the complex scene in the left image in Figure~\ref{fig:case_study2} retains more tokens. This demonstrates how DynTok dynamically compresses low-information-density regions, effectively reducing the number of visual tokens.
For the two rightmost cartoon video frames, ViT struggles with fine lines in the first one, causing coordinate areas to blend into the background, whereas colored regions effectively retain key elements such as the car and trees.

Overall, DynTok achieves reasonable token grouping, effectively initiating new visual token groups for high-information patches while merging tokens that carry similar information.

\section{Conclusion}
In this work, we introduce DynTok, a simple yet effective token compression method that dynamically merges adjacent similar tokens. DynTok significantly enhances the efficiency of video modeling while reducing the computational burden associated with processing long sequences of visual tokens. By retaining only 44\% of the original visual tokens, our approach achieves performance parity with baseline models while substantially improving computational efficiency. Notably, DynTok demonstrates superior scalability with extended input video frames, achieving 65.3\% accuracy on Video-MME and 72.5\% accuracy on MLVU. These results highlight DynTok's ability to not only improve modeling efficiency but also to facilitate easier information extraction for LLMs from lengthy visual token sequences. Overall, DynTok offers a promising approach for designing more efficient and scalable video modeling techniques.

\section*{Limitations}
In principle, this method can be applied to both images and videos. However, since the number of visual tokens in long-video modeling could be huge, we only focus on improving the efficiency and effectiveness of video modeling, and leaves the impact on image understanding as a future work. 
We have not explored the coupling of this token compression method with a series of other methods, such as the scheme of gradually discarding tokens in different layers of the LLMs which is expected to further improve the efficiency. 
\bibliography{acl_latex}

\end{document}

%% file: latex/main_result_table.tex

\begin{table*}[h] 
\centering
\small
\setlength{\tabcolsep}{4pt}
\renewcommand{\arraystretch}{1.2}
\begin{tabular}{@{}l|ccccccc}
\toprule

\multirow{1}{*}{\textbf{Model}} &\multirow{1}{*}{ \textbf{MVbench} }& \multirow{1}{*}{\textbf{PercepTest} }& \multirow{1}{*}{\textbf{NextQA}} &\multirow{1}{*}{\textbf{LongVideo}} & \multirow{1}{*}{\textbf{MLVU}} & \textbf{VideoMME} \\
\midrule
\multicolumn{7}{l}{\textit{Proprietary models}}\\

GPT4o~\cite{openai2024gpt4o} & - & - & - & 66.7 & 64.6 & 71.9  \\
Gemini1.5-pro~\cite{geminiteam2024gemini15unlockingmultimodal} & - & 57.5 & - & 64.0 & - & 75.0  \\
Claude3.5-sonnet~\cite{anthropic2024claude35} & - & - & - & - & - & 60.0  \\
GPT4V~\cite{openai2023gpt4v} & 43.7 & - & - & 61.3 & 49.2 & 59.9  \\
GPT4o-mini~\cite{openai2024gpt4o} & - & - & - & - & - & 64.8  \\

\midrule

\multicolumn{7}{l}{\textit{Open-source models}}\\
Qwen2-VL-7B~\cite{wang2024qwen2vlenhancingvisionlanguagemodels} & 67.0 & 62.3 & - & - & - & 63.3   \\
Intern-VL2.5-8B~\cite{chen2024expanding}  & 72.0& - & - & 68.9 & 60.0 & 64.2   \\
LLaVA-OV-7B~\cite{li2024llavaonevisioneasyvisualtask} & 56.7 & 57.1 & 79.4 & 64.7 & 56.5 & 58.2   \\
LLaVA-Video-7B \cite{zhang2024videoinstructiontuningsynthetic} & 58.6 & 67.9 & 83.2 & 58.2 & 70.8 & 63.3   \\
Dynamic-VLM~\cite{wang2024dynamicvlmsimpledynamicvisual} &-&68.8&-&-&65.0&60.9 \\
DyCoke-7B\cite{Tao2024DyCokeDC}&58.0&	57.6&	78.5&	-&	-&	59.5\\

\midrule

\multicolumn{7}{l}{\textit{Models with comparable training setting}}\\

{Baseline} & 67.9 & 74.0  & 84.1 & 60.1 & 71.0 & 63.6   \\
{DynTok -$zero shot$} & 65.2 & 74.0  & 83.5 & 59.9 & 70.8 & 62.5   \\
{DynTok }&  67.5& 73.5  & 83.7 & 60.4 & 71.3 & 64.0   \\
{DynTok -$more frames$} &-  & - &- &  60.7 &72.5 & 65.3   \\
\bottomrule
\end{tabular}%
\caption{
Performance on various video benchmarks. Results of other models are collected from official reports of the model or the leaderboard of the benchmark. The first three datasets consist of short videos, each with a duration of less than 45 seconds, while the remaining three contain long videos ranging from several minutes to hours.
}
\label{tab:main_result}
\end{table*}

%% file: latex/videomme_category_result_table.tex
\newcommand{\rotationangle}{75}

\begin{table*}[h]
\centering
\small
\renewcommand{\arraystretch}{1.25}
\begin{tabular}{l|c|ccccccc|cccc|c}
\toprule
&&\multicolumn{7}{c|}{\textbf{Perception tasks}} &\multicolumn{4}{c|}{\textbf{Reasoning tasks}}&\multicolumn{1}{c}\textbf{Other}\\ \hline
Model & \rotatebox{\rotationangle}{Overall} & \rotatebox{\rotationangle}{Temporal} & \rotatebox{\rotationangle}{Spatial} & \rotatebox{\rotationangle}{Attribute} & \rotatebox{\rotationangle}{Action} & \rotatebox{\rotationangle}{Object} & \rotatebox{\rotationangle}{OCR} & \rotatebox{\rotationangle}{Counting} & \rotatebox{\rotationangle}{Temporal} & \rotatebox{\rotationangle}{Spatial} & \rotatebox{\rotationangle}{Action} & \rotatebox{\rotationangle}{Object} & \rotatebox{\rotationangle}{Synopsis} \\ \hline

Baseline & 63.6 & 67.3 & 72.2 & 76.6 & 64.5 & 71.5 & 68.4 & 40.3 & 49.2 & 80.4 & 54.7 & 59.5 & 78.6 \\ 
DynTok$zero$ & 62.5 & \underline{\textit{69.1}} & 70.4 & 77.9 & 64.2 & 70.1 & \underline{\textit{64.0}} & 38.4 & 49.7 & 80.4 & \underline{\textit{50.9}} & 59.5 & 77.4 \\ 
DynTok & 64.0 & 74.6 & 70.4 & 77.5 & 62.3 & 70.9 & 69.8 & 41.4 & 51.4 & 78.6 & 54.7 & 61.9 & 77.7 \\ 
\bottomrule
\end{tabular}

\caption{Model performance across different tasks of Video-MME benchmark.}
\label{tab:model_comparison_on_different_tasks}
\end{table*}